\documentclass[11pt,a4paper]{article}
\usepackage[utf8]{inputenc}
\usepackage[]{naaclhlt2019}
\usepackage{times}
\usepackage{latexsym}
\usepackage{graphicx}
\usepackage{booktabs}
\usepackage{url}
\usepackage{multirow}
\usepackage{colortbl}
\usepackage{xcolor}
\usepackage{amsmath,amssymb,bbm}
\usepackage{lipsum}
\usepackage{subcaption}
\usepackage[colorinlistoftodos]{todonotes}
\def\aclpaperid{1458}
\aclfinalcopy

\newcommand\ra{$\rightarrow$}
\newcommand{\MC}[3]{\multicolumn{#1}{#2}{#3}}
\newcommand{\MR}[3]{\multirow{#1}{#2}{#3}}
\newcommand{\vis}{{\texttt{[v]}}}
\newcommand{\B}{\textbf}
\newcommand{\I}{\textit}
\newcommand{\T}{\texttt}
\newcommand{\Dcol}{$\mathcal{D}_{\text{C}}$ }
\newcommand{\Dnou}{$\mathcal{D}_{\text{N}}$ }

\newcommand{\true}[1]{\textbf{#1}}
\newcommand{\false}[1]{\underline{{#1}}}

\title{Probing the Need for Visual Context in Multimodal Machine Translation}

\author{Ozan Caglayan \\
  LIUM, Le Mans University \\
  \texttt{ozan.caglayan@univ-lemans.fr} \\ \And
  Pranava Madhyastha \\
  Imperial College London\\
  \texttt{pranava@imperial.ac.uk} \\ \AND
  Lucia Specia \\
  Imperial College London\\
  \texttt{l.specia@imperial.ac.uk} \\ \And
  Lo\"ic Barrault \\
  LIUM, Le Mans University \\
  \texttt{loic.barrault@univ-lemans.fr}
  }

\date{}

\begin{document}
\maketitle

\begin{abstract}
Current work on multimodal machine translation (MMT) has suggested that the visual modality is either unnecessary or only marginally beneficial. We posit that this is a consequence of the very simple, short and repetitive sentences used in the only available dataset for the task (Multi30K), rendering the source text sufficient as context. In the general case, however, we believe that it is possible to combine visual and textual information in order to ground translations.
In this paper we probe the contribution of the visual modality to state-of-the-art MMT models
by conducting a systematic analysis where we partially deprive the models from source-side textual context.
Our results show that under limited textual context,
models are capable of leveraging the visual input to generate better
translations. This contradicts the current belief that MMT models
disregard the visual modality because of either the quality of the
image features or the way they are integrated into the model.
\end{abstract}
\section{Introduction}
Multimodal Machine Translation (MMT) aims at designing better translation systems which take into account auxiliary inputs such as images. Initially organized as a shared task within the First Conference on Machine Translation (WMT16)~\cite{specia2016shared}, MMT has so far been studied using the Multi30K dataset \cite{Elliott2016}, a multilingual extension of Flickr30K \cite{flickr30k} with translations of the English image descriptions into German, French and Czech \cite{elliott-EtAl:2017:WMT, barrault-EtAl:2018:WMT}.

The three editions of the shared task have seen many exciting approaches that can be broadly categorized as follows: (i) multimodal attention using convolutional features \cite{caglayan2016multiatt,calixto-elliott-frank:2016:WMT,libovicky2017attention,helcl-libovick-varis:2018:WMT} (ii) cross-modal interactions with spatially-unaware global features \cite{calixto2017incorporating,ma-EtAl:2017:WMT1,caglayan2017,madhyastha-wang-specia:2017:WMT} and (iii) the integration of regional features from object detection networks \cite{huang2016attention,grnroos-EtAl:2018:WMT}.
Nevertheless, the conclusion about the contribution of the visual modality is still unclear:
\citet{grnroos-EtAl:2018:WMT} consider their multimodal gains ``modest'' and attribute the largest gain to the usage of external parallel corpora.  \citet{lala-EtAl:2018:WMT} observe that their multimodal word-sense disambiguation approach is not significantly different than the monomodal counterpart. The organizers of the latest edition of the shared task concluded that the multimodal integration schemes explored so far resulted in marginal changes in terms of automatic metrics and human evaluation \cite{barrault-EtAl:2018:WMT}.
In a similar vein, \citet{elliott:2018:EMNLP}  demonstrated that MMT models can translate without significant performance losses even in the presence of features from unrelated images.

These empirical findings seem to indicate that images are ignored by the models and hint at the fact that this is due to representation or modeling limitations. We conjecture that the most plausible reason for the linguistic dominance is that -- at least in Multi30K -- the source text is sufficient to perform the translation, eventually preventing the visual information from intervening in the learning process.
To investigate this hypothesis, we introduce several input degradation regimes (Section~\ref{sec:method}) and revisit state-of-the-art MMT models (Section~\ref{sec:experimental}) to assess their behavior under degraded regimes. We further probe the visual sensitivity by deliberately feeding features from unrelated images. Our results  (Section~\ref{sec:results}) show that MMT models successfully exploit the visual modality when the linguistic context is scarce, but indeed tend to be less sensitive to this modality when exposed to complete sentences.
\section{Input Degradation}
\label{sec:method}

In this section we propose several degradations to the input language modality to simulate conditions where sentences may miss crucial information. We denote a set of translation pairs by $\mathcal{D}$ and indicate degraded variants with subscripts. Both the training and the test sets are degraded.

\paragraph{Color Deprivation.} We consistently replace \I{source} words that refer to colors with a special token \vis\, (\Dcol in Table~\ref{tbl:degraded}). Our hypothesis is that a monomodal system will have to rely on source-side contextual information and biases, while a multimodal architecture could potentially capitalize on color information extracted by exploiting the image and thus obtain better performance. This affects 3.3\% and 3.1\% of the words in the training and the test set, respectively.

\paragraph{Entity Masking.}
The Flickr30K dataset, from which Multi30K is derived, has also been extended with coreference chains to tag mentions of \I{visually depictable} entities in image descriptions \cite{flickr30kent}. We use these to mask out the head nouns in the {\em source} sentences
(\Dnou in Table~\ref{tbl:degraded}). This affects 26.2\% of the words in both the training and the test set. We hypothesize that a multimodal system should heavily rely on the images to infer the missing parts.

\paragraph{Progressive Masking.}
A \I{progressively} degraded variant $\mathcal{D}_k$ replaces all but the first $k$ tokens of \I{source} sentences with \vis\,. Unlike the color deprivation and entity masking, masking out suffixes does not guarantee systematic removal of visual context, but rather simulates an increasingly low-resource scenario.
Overall, we form 16 degraded variants $\mathcal{D}_k$ (Table~\ref{tbl:degraded}) where $k \in \{0,2,\dots,30\}$. We stop at $\mathcal{D}_{30}$ since 99.8\% of the sentences in Multi30K are shorter than 30 words with an average sentence length of 12 words. $\mathcal{D}_0$ -- where the only remaining information is the source sentence length -- is an interesting case from two perspectives: a neural machine translation (NMT) model trained on it resembles a target language model, while an MMT model becomes an image captioner with access to ``expected length information''.

\paragraph{Visual Sensitivity.}
Inspired by \citet{elliott:2018:EMNLP}, we experiment with \I{incongruent decoding} in order to understand how sensitive the multimodal systems are to the visual modality. This is achieved by explicitly violating the test-time semantic congruence across modalities. Specifically, we feed the visual features in \I{reverse} sample order to break image-sentence alignments. Consequently, a model capable of integrating the visual modality would likely deteriorate in terms of metrics.
\begin{table}[t]
\centering
\renewcommand{\arraystretch}{1.0}
\resizebox{.99\columnwidth}{!}{%
\begin{tabular}{@{}lccccccc@{}}
\toprule
$\mathcal{D}$   & \B{a}& \B{lady} & \B{in}   & \B{a}    & \B{blue} & \B{dress} & \B{singing}        \\ \midrule
\Dcol           & a    & lady & in   & a    & \vis & dress & singing        \\
\Dnou           & a    & \vis & in   & a    & blue & \vis  & singing        \\ \midrule
$\mathcal{D}_4$ & a    & lady & in   & a    & \vis & \vis  & \vis           \\
$\mathcal{D}_2$ & a    & lady & \vis & \vis & \vis & \vis  & \vis           \\
$\mathcal{D}_0$ & \vis & \vis & \vis & \vis & \vis & \vis  & \vis           \\
\bottomrule
\end{tabular}%
}
\caption{An example of the proposed input degradation schemes: $\mathcal{D}$ is the original sentence.}
\label{tbl:degraded}
\end{table}
\section{Experimental Setup}
\label{sec:experimental}
\paragraph{Dataset.} We conduct experiments on the English\ra French part of Multi30K. The models are trained on the concatenation of the \I{train} and \I{val} sets (30K sentences) whereas \I{test2016 (dev)} and \I{test2017 (test)} are used for early-stopping and model evaluation, respectively. For \I{entity masking}, we revert to the default Flickr30K splits and perform the model evaluation on \I{test2016}, since \I{test2017} is not annotated for entities.
We use word-level vocabularies of 9,951 English and 11,216 French words.
We use Moses \cite{Moses:2007:acl} scripts to lowercase, normalize and tokenize the sentences with hyphen splitting. The hyphens are stitched back prior to evaluation.

\paragraph{Visual Features.} We use a ResNet-50 CNN \cite{he2016resnet} trained on ImageNet \cite{deng2009imagenet} as image encoder.
Prior to feature extraction, we center and standardize the images using ImageNet statistics, resize the shortest edge to 256 pixels and take a center crop of size 256x256.
We extract spatial features of size 2048x8x8 from the final convolutional layer and apply L$_2$ normalization along the depth dimension \cite{caglayan-EtAl:2018:WMT}. For the non-attentive model, we use the 2048-dimensional global average pooled version (pool5) of the above convolutional features.

\begin{table}[t]
\centering
\renewcommand{\arraystretch}{1.1}
\resizebox{.85\columnwidth}{!}{%
\begin{tabular}{@{}rcccc@{}}
\toprule
            & & $\mathcal{D}$ & \phantom{x}  & $\mathcal{D}_C$ \\
            \cmidrule(l){3-3} \cmidrule(l){5-5}
\phantom{xxxxxxx} NMT &        & 70.6 $\pm$ 0.5 &    & 68.4 $\pm$ 0.1  \\
INIT        && 70.7 $\pm$ 0.2 &    & \B{68.9} $\pm$ 0.1  \\
HIER        && 70.9 $\pm$ 0.3 &    & \B{69.0} $\pm$ 0.3  \\
DIRECT && 70.9 $\pm$ 0.2 & & \B{68.8} $\pm$ 0.3  \\
\bottomrule
\end{tabular}}
\caption{Baseline and color-deprivation METEOR scores: bold systems are significantly different from the NMT system within the \underline{same} column ($p$-value $\le 0.03$).}
\label{tbl:amnmt}
\end{table}

\paragraph{Models.}
Our baseline NMT is an attentive model \cite{Bahdanau2014} with a 2-layer bidirectional GRU encoder \cite{cho2014gru} and a 2-layer conditional GRU decoder \cite{nematus}. The second layer of the decoder receives the output of the attention layer as input.

For the MMT model, we explore the basic multimodal attention (DIRECT) \cite{caglayan2016multiatt} and its hierarchical (HIER) extension \cite{libovicky2017attention}. The former linearly projects the concatenation of textual and visual context vectors to obtain the multimodal context vector, while the latter replaces the concatenation with another attention layer. Finally, we also experiment with encoder-decoder initialization (INIT) \cite{calixto2017incorporating,caglayan2017} where we initialize both the encoder and the decoder using a non-linear transformation of the pool5 features.

\paragraph{Hyperparameters.}
The encoder and decoder GRUs have 400 hidden units
and are initialized with 0 except the multimodal INIT system. All embeddings are 200-dimensional and the decoder embeddings are tied \cite{press2016using}. A dropout of 0.4 and 0.5 is applied on source embeddings and encoder/decoder outputs, respectively \cite{srivastava2014dropout}. The weights are decayed with a factor of $1e\mathrm{-}5$. We use ADAM \cite{kingma2014adam} with a learning rate of $4e\mathrm{-}4$ and mini-batches of 64 samples. The gradients are clipped if the total norm exceeds 1 \cite{pascanu2013difficulty}. The training is early-stopped if \I{dev} set METEOR \cite{meteor} does not improve for ten epochs. All experiments are conducted with \I{nmtpytorch}\footnote{\url{github.com/lium-lst/nmtpytorch}} \cite{nmtpy}.
\section{Results}
\label{sec:results}
We train all systems three times each with different random initialization in order to perform significance testing with \I{multeval} \cite{clark2011better}. Throughout the section, we always report the mean over three runs (and the standard deviation) of the considered metrics. We decode the translations with a beam size of 12.

We first present \I{test2017} METEOR scores for the baseline NMT and MMT systems, when trained on the full dataset $\mathcal{D}$ (Table~\ref{tbl:amnmt}). The first column indicates that, although MMT models perform slightly better on average, they are not significantly better than the baseline NMT. We now introduce and discuss the results obtained under the proposed degradation schemes. Please refer to Table~\ref{tbl:imgcomp} and the appendix for qualitative examples.

\subsection{Color Deprivation}
Unlike the inconclusive results for $\mathcal{D}$,
we observe that all MMT models are significantly better than NMT when color deprivation is applied (\Dcol in Table~\ref{tbl:amnmt}). If we further focus on the subset of the test set subjected to color deprivation (247 sentences), the gain increases to 1.6 METEOR for HIER. For the latter subset, we also computed the average color accuracy per sentence and found that the attentive models are 12\% better than the NMT (32.5\ra44.5) whereas the INIT model only brings 4\% (32.5\ra36.5) improvement. This shows that more complex MMT models are better at integrating visual information to perform better.
\begin{figure}[t]
\centering
\includegraphics[width=.95\columnwidth]{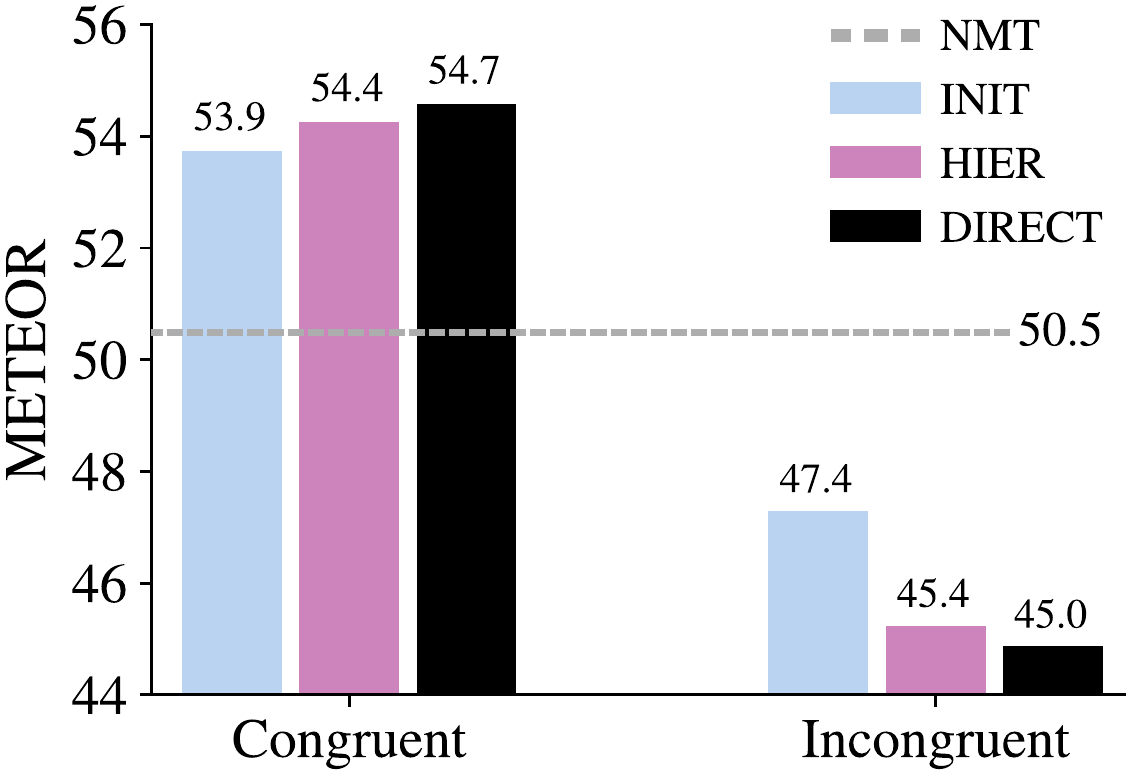}
\caption{Entity masking: all masked MMT models are significantly better than the masked NMT (dashed). Incongruent decoding severely worsens all systems.
The vanilla NMT baseline is 75.9\protect\footnotemark .}
\label{fig:entity}
\end{figure}

\footnotetext{Since entity masking uses Flickr30K splits (Section ~\ref{sec:experimental}) rather than our splits, the scores are not comparable to those from other experiments in this paper.}

\subsection{Entity Masking}
\begin{figure}[t]
\centering
\includegraphics[width=.83\columnwidth]{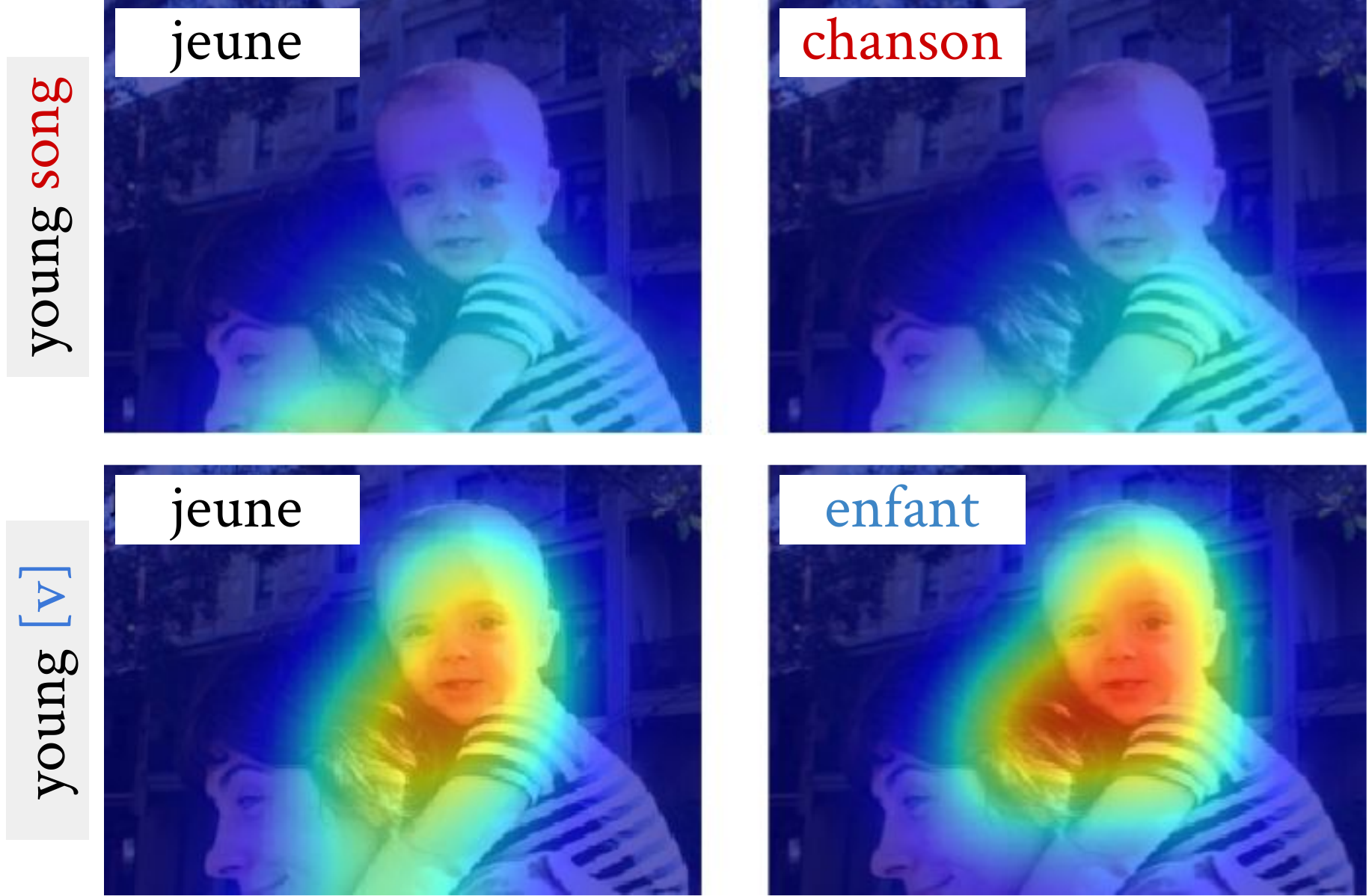}
\caption{Baseline MMT (top) translates the misspelled ``son'' while the masked MMT (bottom) correctly produces ``enfant'' (child) by focusing on the image.}
\label{fig:ex_att}
\end{figure}

The gains are much more prominent with entity masking, where the degradation occurs at a larger scale: Attentive MMT models show up to 4.2 METEOR improvement over NMT (Figure~\ref{fig:entity}).
We observed a large performance drop with \I{incongruent decoding}, suggesting that the visual modality is now much more important than previously demonstrated \cite{elliott:2018:EMNLP}.
A comparison of attention maps produced by the baseline and masked MMT models
reveals that the attention weights are more consistent in the latter. An interesting example is given in Figure~\ref{fig:ex_att} where the masked MMT model attends to the correct region of the image and successfully translates a dropped word that was otherwise a spelling mistake (``son''\ra ``son\B{g}'').
\begin{table}[t!]
\centering
\renewcommand{\arraystretch}{1.1}
\resizebox{.96\columnwidth}{!}{%
\begin{tabular}{@{}rccc@{}}
\toprule
          & \MC{3}{c}{$+$ Gain ($\downarrow$ Incongruence Drop)} \\ \cmidrule(l){2-4}
          & INIT     & HIER      & DIRECT   \\ \cmidrule(l){2-2} \cmidrule(l){3-3} \cmidrule(l){4-4}
Czech     & $+$1.4 ($\downarrow$ 2.9) & $+$1.7 ($\downarrow$ 3.5) & $+$1.7 ($\downarrow$ 4.1)    \\ 
German    & $+$2.1 ($\downarrow$ 4.7) & $+$2.5 ($\downarrow$ 5.9) & $+$2.7 ($\downarrow$ 6.5)    \\ 
French    & $+$3.4 ($\downarrow$ 6.5) & $+$3.9 ($\downarrow$ 9.0) & $+$4.2 ($\downarrow$ 9.7)    \\ 
\bottomrule
\end{tabular}}
\caption{\I{Entity masking} results across three languages: all MMT models perform significantly better than their NMT counterparts ($p$-value $\leq 0.01$). The incongruence drop applies on top of the MMT score.}
\label{tbl:entity_all}
\end{table}

\paragraph{Czech and German.}
In order to understand whether the above observations are also consistent across different languages, we extend the \I{entity masking} experiments to German and Czech parts of Multi30K. Table~\ref{tbl:entity_all} shows the gain of each MMT system with respect to the NMT model and the subsequent drop caused by incongruent decoding\footnote{For example, the INIT system for French (Figure~\ref{fig:entity}) surpasses the baseline (50.5) by reaching 53.9 (+3.4), which ends up at 47.4 ($\downarrow$ 6.5) after incongruent decoding.}.
First, we see that the multimodal benefits clearly hold for German and Czech, although the gains are lower than for French\footnote{This is probably due to the morphological richness of DE and CS which is suboptimally handled by word-level MT.}.
Second, when we compute the average drop from using incongruent images across all languages, we see how conservative the INIT system is ($\downarrow$ 4.7) compared to HIER ($\downarrow$ 6.1) and DIRECT ($\downarrow$ 6.8).
This raises a follow-up question as to whether the hidden state initialization eventually loses its impact throughout the recurrence where, as a consequence, the only modality processed is the text.

\begin{figure}[t!]
\centering
\includegraphics[width=1\columnwidth]{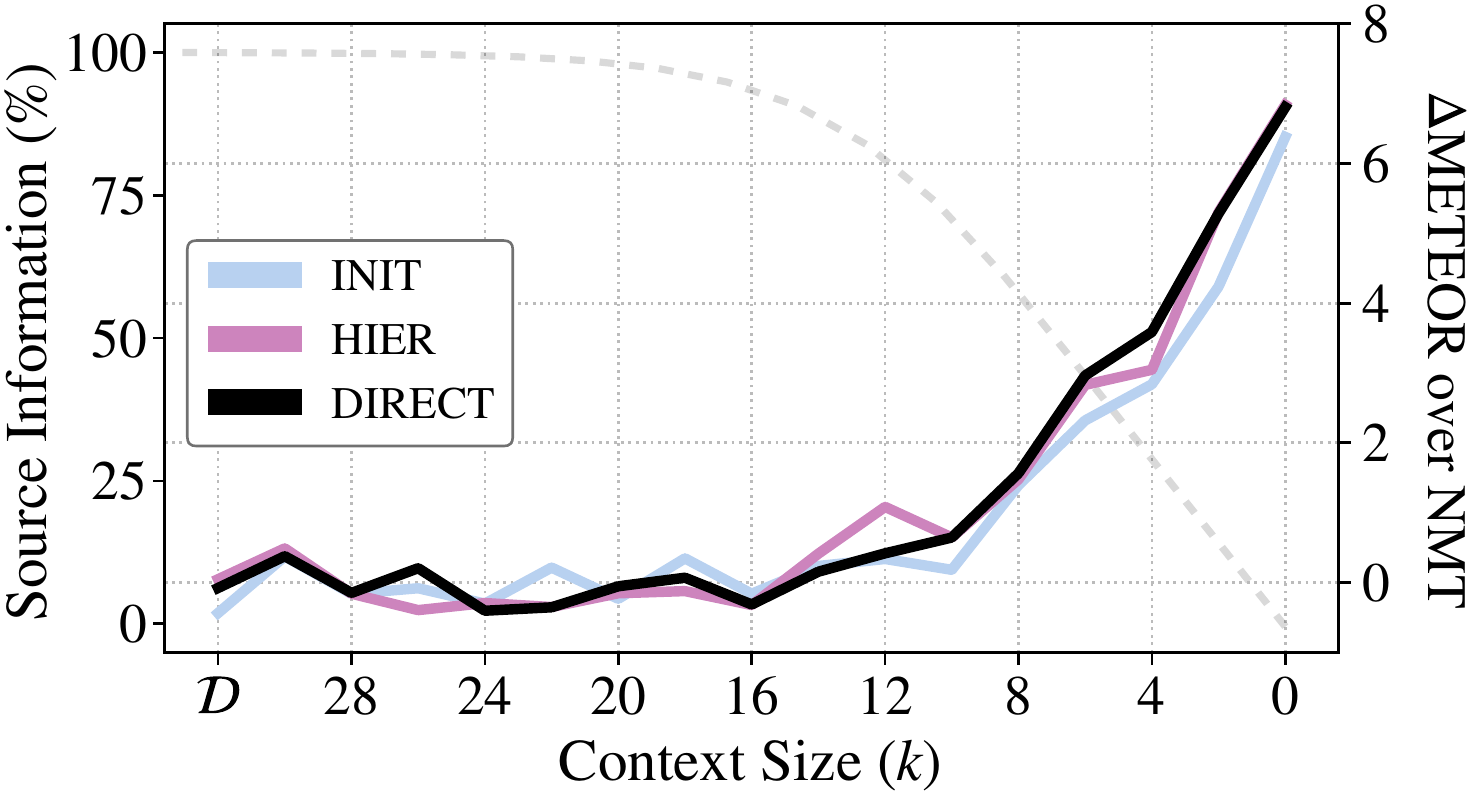}
\caption{Multimodal gain in absolute METEOR for \I{progressive masking}: the dashed gray curve indicates the percentage of non-masked words in the training set.}
\label{fig:amnmtovernmt}
\end{figure}

\begin{table}[t!]
\centering
\renewcommand{\arraystretch}{1.07}
\resizebox{1\columnwidth}{!}{%
\begin{tabular}{@{}lrrrrr@{}}
\toprule
 & $\mathcal{D}_4$ & $\mathcal{D}_6$ & $\mathcal{D}_{12}$ & $\mathcal{D}_{20}$ & $\mathcal{D}$\phantom{x} \\ \cmidrule(l){2-2} \cmidrule(l){3-3} \cmidrule(l){4-4} \cmidrule(l){5-5} \cmidrule(l){6-6}
DIRECT                  & \B{32.3}      & \B{42.2}      & \B{64.5}      & \B{70.1}      & \B{70.9}       \\
Incongruent Dec.    & $\downarrow$ 6.4      & $\downarrow$ 5.5      & $\downarrow$ 1.4      & $\downarrow$ 0.7      & $\downarrow$ 0.7       \\ \midrule
Blinding                & $\downarrow$ 3.9      & $\downarrow$ 2.9      & $\downarrow$ 0.4      & $\downarrow$ 0.5      & $\downarrow$ 0.3        \\ 
NMT                     & $\downarrow$ 3.7      & $\downarrow$ 2.6      & $\downarrow$ 0.6      & $\downarrow$ 0.2      & $\downarrow$ 0.3        \\  
\bottomrule
\end{tabular}}
\caption{The impact of incongruent decoding for \I{progressive masking}: all METEOR differences are against the DIRECT model. The blinded systems are both trained and decoded using incongruent features.}
\label{tbl:visdeg}
\end{table}

\subsection{Progressive Masking}
\begin{table*}[htbp!]
\renewcommand{\arraystretch}{1.0}
\centering
\resizebox{.95\textwidth}{!}{%
\begin{tabular}{cl@{}}
\toprule
\MR{7}{*}{\includegraphics[height=2.5cm]{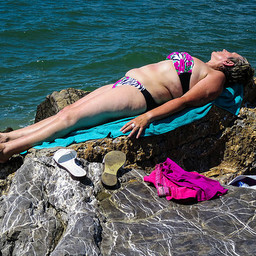}} &
\T{SRC:} an older woman in \vis \vis \vis \vis \vis \vis \vis \vis \vis \vis \vis \\
& \T{NMT:} une femme âgée avec un \false{t-shirt blanc} et des lunettes de soleil est assise sur un \false{banc} \\
& \phantom{\T{NMT:}} \I{(an older woman with a white t-shirt and sunglasses is sitting on a bank)} \\
& \T{MMT:} une femme âgée en \true{maillot de bain rose} est assise sur un \true{rocher au bord de l'eau} \\
& \phantom{\T{MMT:}} \I{(an older woman with a pink swimsuit is sitting on a rock at the seaside)} \\
& \T{REF:} une femme âgée \true{en bikini} bronze sur \true{un rocher au bord de l'océan} \\
& \phantom{\T{REF:}} \I{(an older woman in bikini is tanning on a rock at the edge of the ocean)} \\
\midrule
\MR{6}{*}{\includegraphics[height=2.5cm]{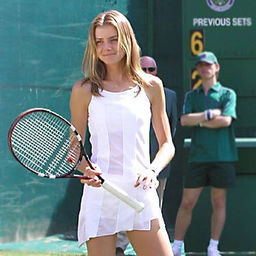}} &
\T{SRC:} a young \vis\,in \vis\,holding a tennis \vis \\
& \T{NMT:} \false{un} jeune \false{garçon} en \false{bleu} tenant une raquette de tennis \\
& \phantom{\T{NMT:}} \I{(a young boy in blue holding a tennis racket)} \\
& \T{MMT:} \true{une} jeune \true{femme} en \true{blanc} tenant une raquette de tennis \\
& \T{REF:} \true{une} jeune \true{femme} en \true{blanc} tenant une raquette de tennis \\
& \phantom{\T{MMT:}} \I{(a young girl in white holding a tennis racket)} \\
\midrule
\MR{6}{*}{\includegraphics[height=2.5cm]{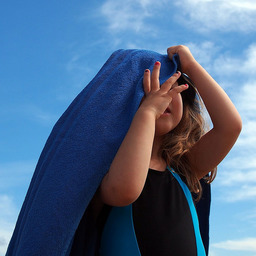}} &
\T{SRC:} little girl covering her face with a \vis\,towel \\
& \T{NMT:} une petite fille couvrant son visage avec une serviette \false{blanche}  \\
& \phantom{\T{NMT:}} \I{(a little girl covering her face with a white towel)}  \\
& \T{MMT:} une petite fille couvrant son visage avec une serviette \true{bleue}  \\
& \T{REF:} une petite fille couvrant son visage avec une serviette \true{bleue}  \\
& \phantom{\T{MMT:}} \I{(a little girl covering her face with a blue towel)}  \\
\bottomrule
\end{tabular}}
\caption{Qualitative examples from progressive masking, entity masking and color deprivation, respectively. Underlined and bold words highlight the bad and good lexical choices. MMT is an attentive system.}
\label{tbl:imgcomp}
\end{table*}

Finally, we discuss the results of the progressive masking experiments for French. Figure~\ref{fig:amnmtovernmt} clearly shows that as the sentences are progressively degraded, all MMT systems are able to leverage the visual modality. When the multimodal task becomes image captioning at $k\mathrm{=}0$, MMT models improve over the language-model counterpart by $\sim$7 METEOR. Further qualitative examples show that the systems perform surprisingly well by producing visually plausible sentences (see Table~\ref{tbl:imgcomp} and the Appendix).


To get a sense of the visual sensitivity, we pick the DIRECT models trained on four degraded variants and perform \I{incongruent decoding}. We notice that as the amount of linguistic information increases, the gap narrows down: the MMT system gradually becomes less perplexed by the incongruence or, put in other words, less sensitive to the visual modality (Table~\ref{tbl:visdeg}).\newpage
\noindent We then conduct a contrastive ``blinding'' experiment where the DIRECT models are not only fed with incongruent features at decoding time but also trained with them from scratch. The results suggest that the blinded models learn to ignore the visual modality. In fact, their performance is equivalent to NMT models.
\section{Discussion and Conclusions}

We presented an in-depth study on the potential contribution of images for multimodal machine translation. Specifically, we analysed the behavior of state-of-the-art MMT models under several degradation schemes in the Multi30K dataset, in order to reveal and understand the impact of textual predominance.
Our results show that the models explored are able to integrate the visual modality if the available modalities are complementary rather than redundant. In the latter case, the primary modality (text) sufficient to accomplish the task.
This dominance effect corroborates the seminal work of \citet{colavita1974} in Psychophysics where it has been demonstrated that visual stimuli dominate over the auditory stimuli when humans are asked to perform a simple audiovisual discrimination task.
Our investigation using source degradation also suggests that visual grounding can increase the robustness of machine translation systems by mitigating input noise such as errors in the source text.
In the future, we would like to devise models that can learn when and how to integrate multiple modalities by taking care of the complementary and redundant aspects of them in an intelligent way.
\section*{Acknowledgments}
This work is a follow-up on the research efforts conducted within the ``Grounded sequence-to-sequence transduction'' team of the JSALT 2018 Workshop. We would like to thank Jindřich Libovický for contributing the hierarchical attention to \I{nmtpytorch} during the workshop. We also thank the reviewers for their valuable comments.

Ozan Caglayan and Lo\"ic Barrault received funding from the French National Research Agency (ANR) through the CHIST-ERA M2CR project under the contract ANR-15-CHR2-0006-01. Lucia Specia and Pranava Madhyastha received funding from the MultiMT (H2020 ERC Starting Grant No. 678017) and MMVC (Newton Fund Institutional Links Grant, ID 352343575) projects.

\bibliographystyle{acl_natbib}
\bibliography{naaclhlt2019}

\appendix
\section{Qualitative Examples}
\begin{figure*}[t]
\centering
\begin{subfigure}[b]{0.99\textwidth}
\includegraphics[width=.98\textwidth]{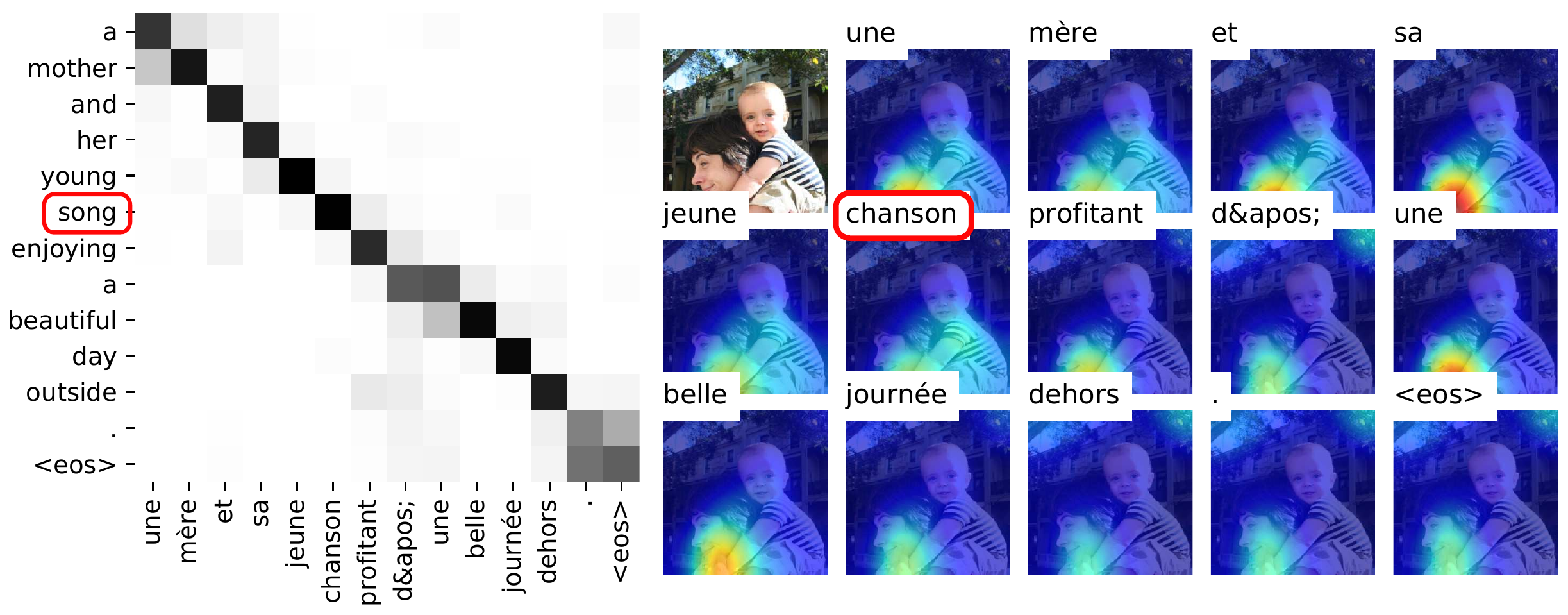}
\caption{Baseline (non-masked) MMT}
\label{fig:app_att_1_1}
\end{subfigure}
\begin{subfigure}[b]{0.99\textwidth}
\includegraphics[width=.98\textwidth]{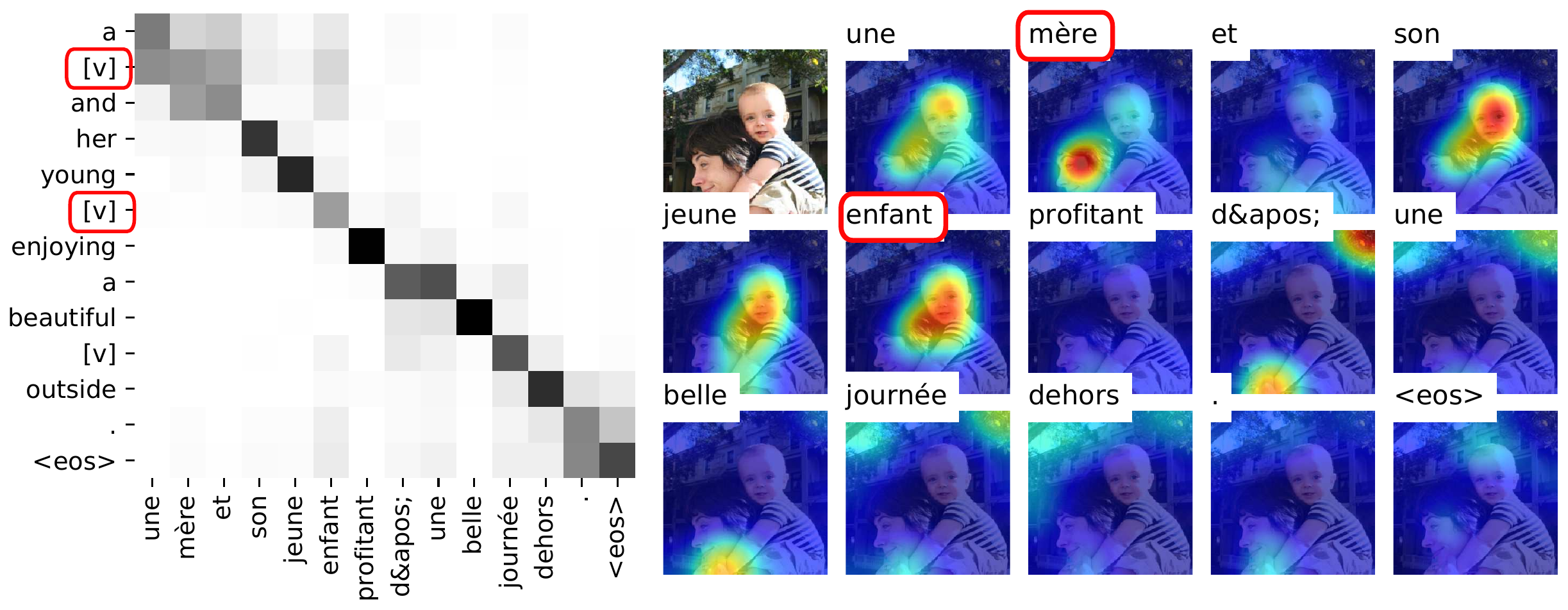}
\caption{Entity-masked MMT}
\label{fig:app_att_1_2}
\end{subfigure}
\caption{Attention example from entity masking experiments: (a) Baseline MMT translates the misspelled ``son'' (song \ra\,chanson) while (b) the masked MMT achieves a correct translation (\vis\ra\,enfant) by exploiting the visual modality.}
\label{fig:app_att_1}
\end{figure*}

\begin{figure*}[t]
\centering
\begin{subfigure}[b]{0.99\textwidth}
\includegraphics[width=.98\textwidth]{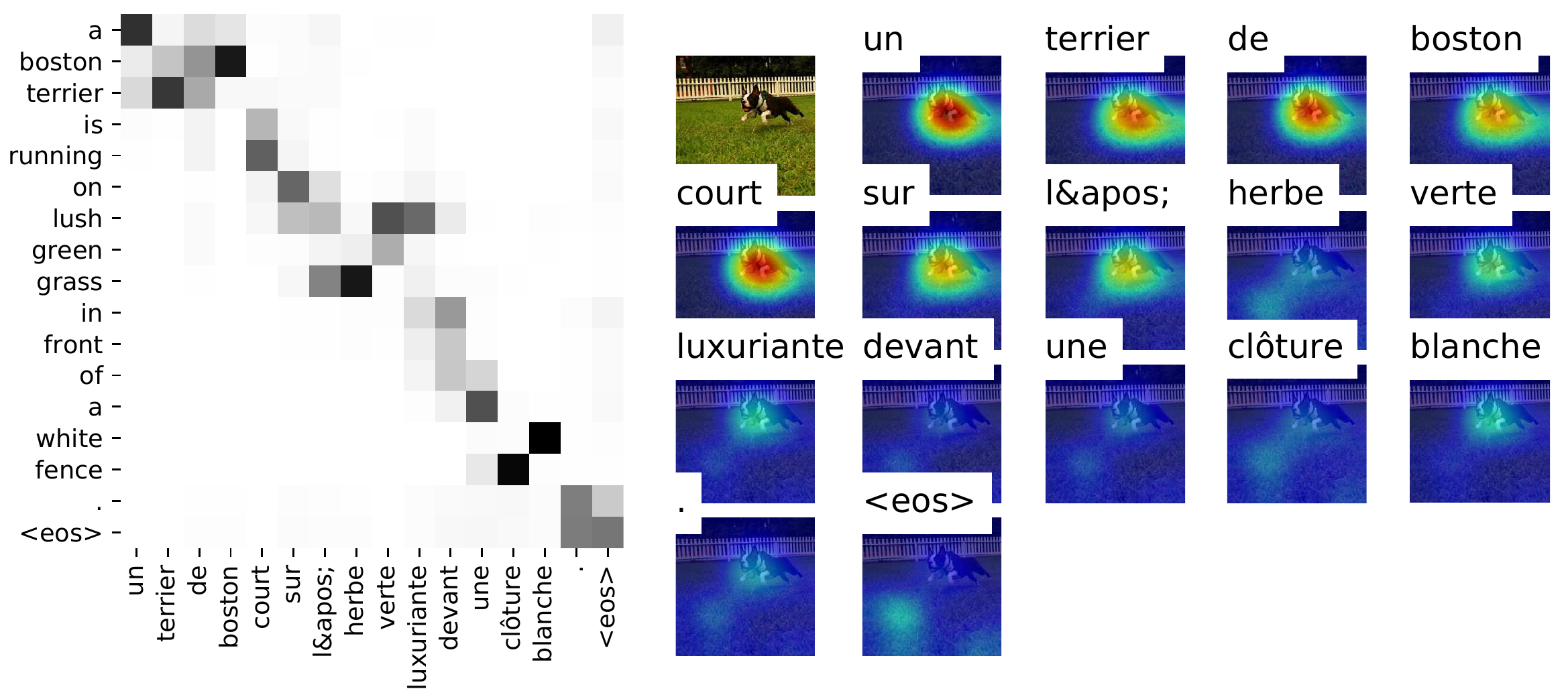}
\caption{Baseline (non-masked) MMT}
\label{fig:app_att_2_1}
\end{subfigure}
\begin{subfigure}[b]{0.99\textwidth}
\includegraphics[width=.98\textwidth]{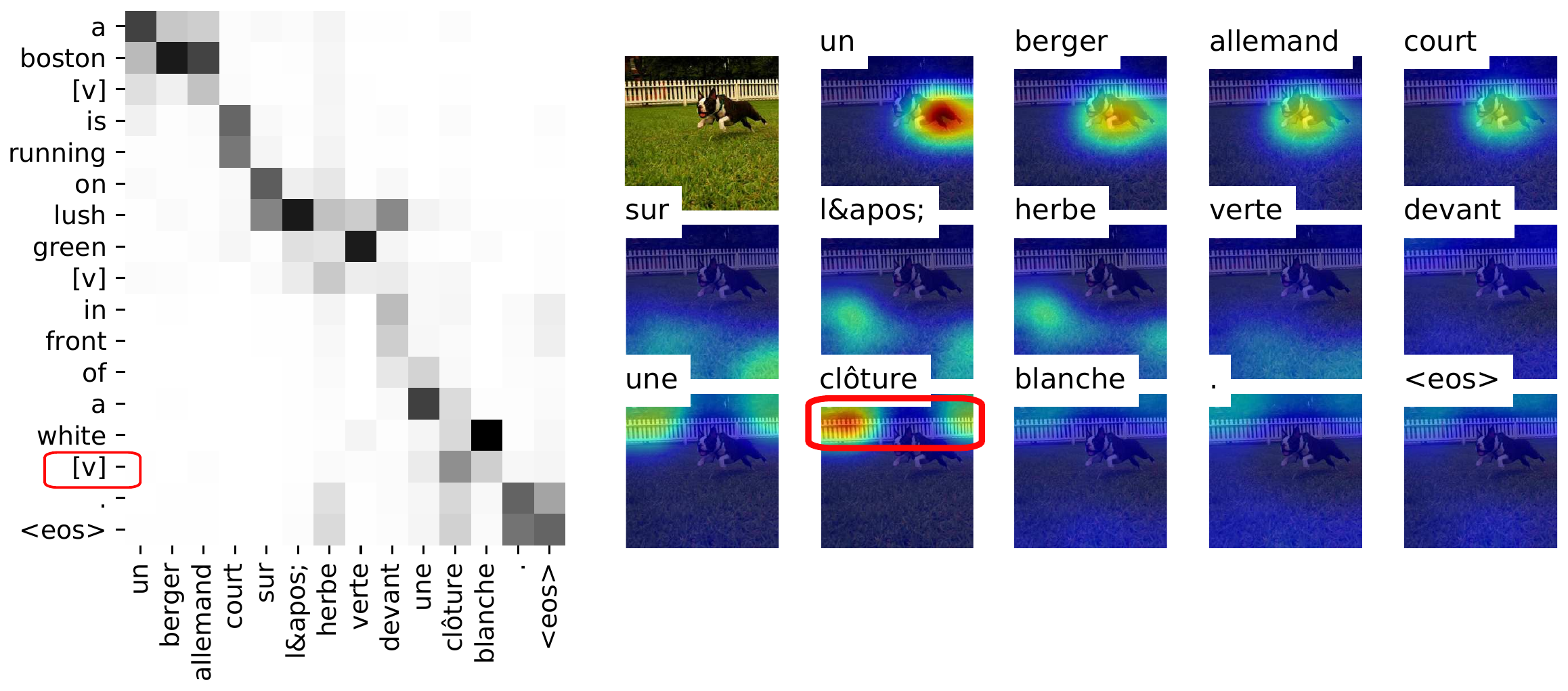}
\caption{Entity-masked MMT}
\label{fig:app_att_2_2}
\end{subfigure}
  \caption{Attention example from entity masking experiments where \I{terrier, grass} and \I{fence} are dropped from the source sentence: (a) Baseline MMT is not able to shift attention from the salient \I{dog} to the \I{grass} and \I{fence}, (b) the attention produced by the masked MMT first shifts to the background area while translating ``on lush green \vis'' then focuses on the \I{fence}.}
\label{fig:app_att_2}
\end{figure*}

In this appendix, we provide further translation examples for color deprivation (Table~\ref{tbl:app_color}),
entity masking (Table~\ref{tbl:app_entity}) and progressive masking (Table~\ref{tbl:app_pro}).
Specifically for
the entity masking experiments, we also give further examples to showcase the
behavior of the visual attention in Figure~\ref{fig:app_att_1} and Figure~\ref{fig:app_att_2}.

\vspace*{9em}
\begin{table*}[h]
\renewcommand{\arraystretch}{1}
\centering
\resizebox{.9\textwidth}{!}{%
\begin{tabular}{cl@{}}
\toprule
\MR{7}{*}{\includegraphics[height=2.9cm]{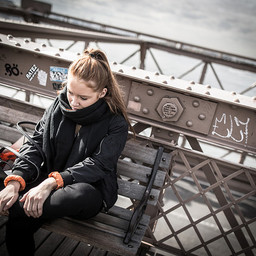}} &
\\
& \T{SRC:\phantom{...}} a girl in \vis\,is sitting on a bench \\
& \T{NMT:\phantom{...}} pink \\
& \T{Init:\phantom{..}} pink \\
& \T{Hier:\phantom{..}} \true{black} \\
& \T{Direct:} \true{black} \\
\\
\midrule
\MR{7}{*}{\includegraphics[height=2.9cm]{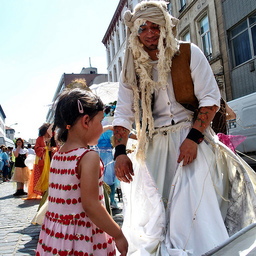}} &
\\
& \T{SRC:\phantom{...}} a man dressed in \vis\,talking to a girl \\
& \T{NMT:\phantom{...}} black \\
& \T{Init:\phantom{..}} black \\
& \T{Hier:\phantom{..}} \true{white} \\
& \T{Direct:} \true{white} \\
\\
\midrule
\MR{7}{*}{\includegraphics[height=2.9cm]{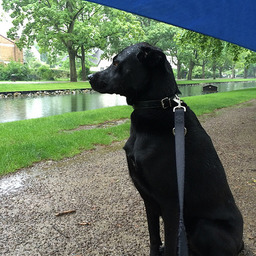}} &
\\
& \T{SRC:\phantom{...}} a \vis\,dog sits under a \vis\,umbrella \\
& \T{NMT:\phantom{...}} brown / \true{blue} \\
& \T{Init:\phantom{..}} \true{black} / \true{blue} \\
& \T{Hier:\phantom{..}} \true{black} / \true{blue} \\
& \T{Direct:} \true{black} / \true{blue} \\
\\
\midrule
\MR{7}{*}{\includegraphics[height=2.9cm]{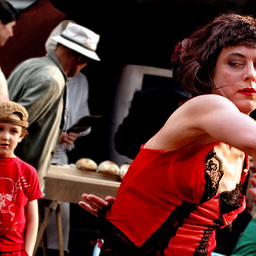}} &
\\
& \T{SRC:\phantom{...}} a woman in a \vis\,top is dancing as a woman and boy in a \vis\,shirt watch \\
& \T{NMT:\phantom{...}} blue / blue \\
& \T{Init:\phantom{..}} blue / blue \\
& \T{Hier:\phantom{..}} \true{red} / \true{red} \\
& \T{Direct:} \true{red} / \true{red} \\
\\
\midrule
\MR{7}{*}{\includegraphics[height=2.9cm]{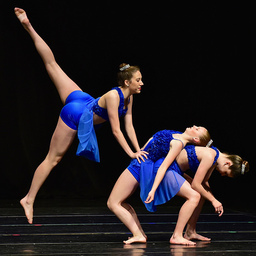}} &
\\
& \T{SRC:\phantom{...}} three female dancers in \vis\,dresses are performing a dance routine \\
& \T{NMT:\phantom{...}} white \\
& \T{Init:\phantom{..}} white \\
& \T{Hier:\phantom{..}} white \\
& \T{Direct:} \true{blue} \\
\\
\bottomrule
\end{tabular}}
\caption{Color deprivation examples from the English$\rightarrow$French models: bold indicates correctly predicted cases. The colors generated by the models are shown in English for the sake of clarity.}
\label{tbl:app_color}
\end{table*}

\begin{table*}[t]
\renewcommand{\arraystretch}{1.0}
\centering
\resizebox{.8\textwidth}{!}{%
\begin{tabular}{cl@{}}
\toprule
\MR{6}{*}{\includegraphics[height=2.5cm]{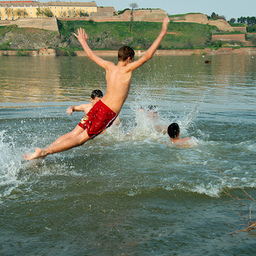}} &
\T{SRC:} a \vis\,in a red \vis\,plays in the \vis \\
& \T{NMT:} un garçon en \false{t-shirt} rouge joue dans la \false{neige} \\
& \phantom{\T{NMT:}} \I{(a boy in a red t-shirt plays in the snow)} \\
& \T{MMT:} un garçon en \true{maillot de bain} rouge joue dans \true{l'eau} \\
& \T{REF:} un garçon en \true{maillot de bain} rouge joue dans \true{l'eau} \\
& \phantom{\T{MMT:}} \I{(a boy in a red swimsuit plays in the water)} \\
\midrule
\MR{6}{*}{\includegraphics[height=2.5cm]{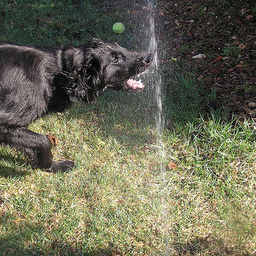}} &
\T{SRC:} a \vis\,drinks \vis\,outside on the \vis \\
& \T{NMT:} un \false{homme} boit du \false{vin} dehors sur le \false{trottoir} \\
& \phantom{\T{NMT:}} \I{(a man drinks wine outside on the sidewalk)} \\
& \T{MMT:} un \true{chien} boit de \true{l'eau} dehors sur \true{l'herbe} \\
& \T{REF:} un \true{chien} boit de \true{l'eau} dehors sur \true{l'herbe} \\
& \phantom{\T{MMT:}} \I{(a dog drinks water outside on the grass)} \\
\midrule
\MR{6}{*}{\includegraphics[height=2.5cm]{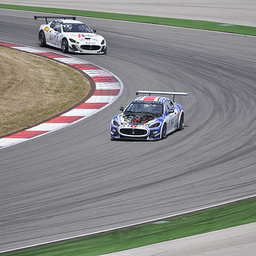}} &
\T{SRC:} two \vis\,are driving on a \vis \\
& \T{NMT:} deux \false{hommes} font du \false{vélo} sur une route \\
& \phantom{\T{NMT:}} \I{(two men riding bicycles on a road)} \\
& \T{MMT:} deux \true{voitures roulent sur une piste} \\
& \phantom{\T{MMT:}} \I{(two cars driving on a track/circuit)} \\
& \T{REF:} deux \true{voitures roulent sur} un circuit \\
\midrule
\MR{7}{*}{\includegraphics[height=2.5cm]{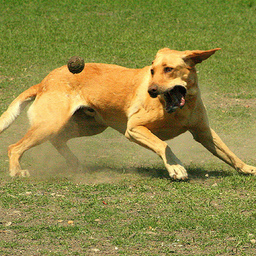}} &
\T{SRC:} a \vis\,turns on the \vis\,to pursue a flying \vis \\
& \T{NMT:} un \false{homme} tourne sur la \false{plage} pour attraper un frisbee volant \\
& \phantom{\T{NMT:}} \I{(a man turns on the beach to catch a flying frisbee)} \\
& \T{MMT:} un \true{chien} tourne sur \true{l'herbe} pour attraper un frisbee volant \\
& \phantom{\T{MMT:}} \I{(a dog turns on the grass to catch a flying frisbee)} \\
& \T{REF:} un \true{chien} tourne sur \true{l'herbe} pour poursuivre une balle en l'air \\
& \phantom{\T{MMT:}} \I{(a dog turns on the grass to chase a ball in the air)} \\
\midrule
\MR{6}{*}{\includegraphics[height=2.5cm]{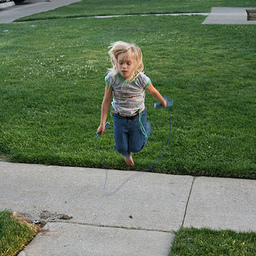}} &
\T{SRC:} a \vis\,jumping \vis\,on a \vis\,near a parking \vis \\
& \T{NMT:} un \false{homme} sautant à \false{cheval} sur une \false{plage} près d'un parking \\
& \phantom{\T{NMT:}} \I{(a man jumping on a beach near a parking lot)} \\
& \T{MMT:} une \true{fille} sautant à la \true{corde} sur un \true{trottoir} près d'un parking \\
& \T{REF:} une \true{fille} sautant à la \true{corde} sur un \true{trottoir} près d'un parking \\
& \phantom{\T{REF:}} \I{(a girl jumping rope on a sidewalk near a parking lot)} \\
\bottomrule
\end{tabular}}
\caption{Entity masking examples from the English$\rightarrow$French models: underlined and bold words highlight bad and good lexical choices, respectively. English translations are provided in parentheses. MMT is an attentive model.}
\label{tbl:app_entity}
\end{table*}


\begin{table*}[t]
\renewcommand{\arraystretch}{1.0}
\centering
\resizebox{.99\textwidth}{!}{%
\begin{tabular}{cl@{}}
\toprule
\MR{7}{*}{\includegraphics[height=2.6cm]{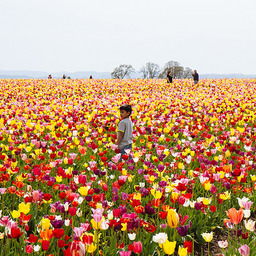}} &
\T{SRC:} a child \vis \vis \vis \vis \vis \vis \\
& \T{NMT:} un enfant \false{avec des lunettes de soleil en train de jouer au tennis} \\
& \phantom{\T{NMT:}} \I{(a child with sunglasses playing tennis)} \\
& \T{MMT:} un enfant \true{est debout dans un champ de fleurs} \\
& \phantom{\T{MMT:}} \I{(a child is standing in field of flowers)}\\
& \T{REF:} un enfant \true{dans un champ de tulipes} \\
& \phantom{\T{REF:}} \I{(a child in a field of tulips)}\\
\midrule
\MR{6}{*}{\includegraphics[height=2.6cm]{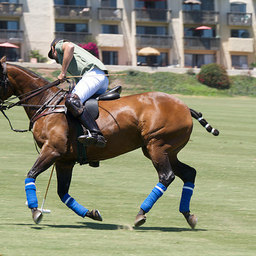}} &
\T{SRC:} a jockey riding his \vis \vis \\
& \T{NMT:} un jockey sur son \false{vélo} \\
& \phantom{\T{NMT:}} \I{(a jockey on his bike)} \\
& \T{MMT:} un jockey sur son \true{cheval} \\
& \T{REF:} un jockey sur son \true{cheval} \\
& \phantom{\T{MMT:}} \I{(a jockey on his horse)} \\
\midrule
\MR{6}{*}{\includegraphics[height=2.6cm]{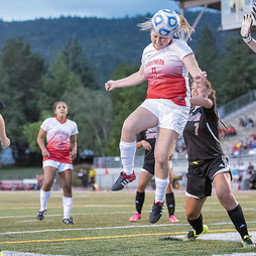}} &
\T{SRC:} girls are playing a \vis \vis \vis \\
& \T{NMT:} des filles jouent à un \false{jeu de cartes} \\
& \phantom{\T{NMT:}} \I{(girls are playing a card game)} \\
& \T{MMT:} des filles jouent un \true{match de football} \\
& \T{REF:} des filles jouent un \true{match de football} \\
& \phantom{\T{MMT:}} \I{(girls are playing a football match)} \\
\midrule
\MR{7}{*}{\includegraphics[height=2.6cm]{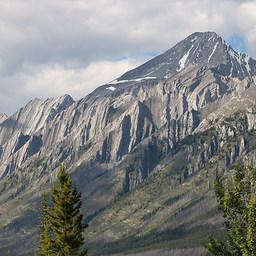}} &
\T{SRC:} trees are in front \vis \vis \vis \vis \vis \\
& \T{NMT:} des \false{vélos} sont devant un \false{bâtiment} en plein air \\
& \phantom{\T{NMT:}} \I{(bicycles are in front of an outdoor building)} \\
& \T{MMT:} des \true{arbres} sont devant la \true{montagne} \\
& \phantom{\T{MMT:}} \I{(trees are in front of the mountain)} \\
& \T{REF:} des \true{arbres} sont devant une grande \true{montagne} \\
& \phantom{\T{REF:}} \I{(trees are in front of a big mountain)} \\
\midrule
\MR{7}{*}{\includegraphics[height=2.6cm]{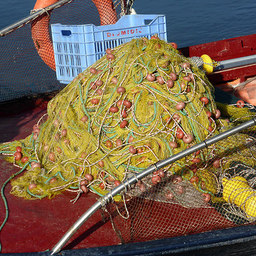}} &
\T{SRC:} a fishing net on the deck of a \vis \vis \\
& \T{NMT:} un filet de pêche sur la \false{terrasse d'un bâtiment} \\
& \phantom{\T{NMT:}} \I{(a fishing net on the terrace of a building)} \\
& \T{MMT:} un filet de pêche sur le \true{pont d'un bateau} \\
& \phantom{\T{MMT:}} \I{(a fishing net on the deck of a boat)} \\
& \T{REF:} un filet de pêche sur le \true{pont d'un bateau} rouge \\
& \phantom{\T{REF:}} \I{(a fishing net on the deck of a red boat)} \\
\midrule
\MR{7}{*}{\includegraphics[height=2.6cm]{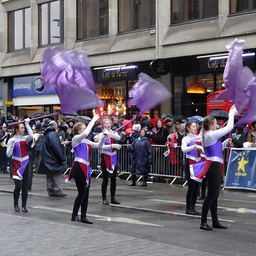}} &
\T{SRC:} girls wave purple flags \vis \vis \vis \vis \vis \vis \vis \\
& \T{NMT:} des filles en t-shirts violets sont \false{assises sur des chaises dans une salle de classe} \\
& \phantom{\T{NMT:}} \I{(girls in purple t-shirts are sitting on chairs in a classroom)} \\
& \T{MMT:} des filles en costumes violets \true{dansent dans une rue en ville} \\
& \phantom{\T{MMT:}} \I{(girls in purple costumes dance on a city street)} \\
& \T{REF:} des filles agitent des drapeaux violets tandis qu'elles défilent dans la rue \\
& \phantom{\T{REF:}} \I{(girls wave purple flags as they parade down the street)} \\
\bottomrule
\end{tabular}}
\caption{English$\rightarrow$French progressive masking examples: underlined and bold words highlight bad and good lexical choices, respectively. English translations are provided in parentheses. MMT is an attentive model.}
\label{tbl:app_pro}
\end{table*}

\end{document}